\def\year{2020}\relax
\documentclass[letterpaper]{article} 
\usepackage{aaai20}  
\usepackage{times}  
\usepackage{helvet} 
\usepackage{courier}  
\usepackage[hyphens]{url}  
\usepackage{graphicx} 
\usepackage{graphicx}  
\usepackage{textcomp}
\usepackage{amsmath,amssymb,amsfonts}
\usepackage{xcolor}
\usepackage{algorithm,algpseudocode}  
\usepackage{algorithmicx} 
\usepackage{thmtools}
\usepackage{subfigure}
\usepackage{rotating}
\usepackage{type1cm}

\title{Outlier Detection Ensemble with Embedded Feature Selection}

\author{{Li Cheng\textsuperscript{\rm 1}, Yijie Wang\textsuperscript{\rm 1}\thanks{Corresponding author}, Xinwang Liu, Bin Li\textsuperscript{\rm 1}}\\ 
\textsuperscript{\rm 1}Science and Technology on Parallel and Distributed Processing Laboratory\\ 
College of Computer, National University of Defense Technology\\
Changsha, China \\
{\{chengli09, wangyijie, liuxinwang, libin16a\}@nudt.edu.cn} 
}

\begin{document}

\maketitle

\begin{abstract}
Feature selection places an important role in improving the performance of outlier detection, especially for noisy data. 
Existing methods usually perform feature selection and outlier scoring separately, which would select feature subsets that may not optimally serve for outlier detection, leading to unsatisfying performance.
In this paper, we propose an outlier detection ensemble framework with embedded feature selection (ODEFS), to address this issue. 
Specifically, for each random sub-sampling based learning component, ODEFS unifies feature selection and outlier detection into a pairwise ranking formulation to learn feature subsets that are tailored for the outlier detection method. 
Moreover, we adopt the thresholded self-paced learning to simultaneously optimize feature selection and example selection, which is helpful to improve the reliability of the training set. 
After that, we design an alternate algorithm with proved convergence to solve the resultant optimization problem. 
In addition, we analyze the generalization error bound of the proposed framework, which provides theoretical guarantee on the method and insightful practical guidance. 
Comprehensive experimental results on 12 real-world datasets from diverse domains validate the superiority of the proposed ODEFS.
\end{abstract}

\section{Introduction}
Outlier detection has been intensively studied and widely used in various applications, such as medical diagnosis \cite{wang2019study}, fraud detection \cite{wang2018serendipitous,wang2013survey}, and information security \cite{kang2019similarity}, to name just a few. 
In such real-world applications, it is not uncommon to see that there are many irrelevant or redundant features among data when performing outlier detection \cite{kang2019robust,liu2007computational}. 
It has been shown that the performance of outlier detection can be significantly improved by only using the informative feature subsets \cite{pang2018sparse,Keller2012HiCS}.
Therefore, feature/subspace selection, which can help to remove noisy features, places an important role in improving the performance of outlier detection, especially for noisy data.

Feature selection based outlier detection methods select relevant feature subsets for the subsequent outlier detection method, with the aim to alleviate the negative effect brought by noisy features.
Many works on this regard have been developed. 
Early attempts often separate the feature searching from the subsequent outlier scoring methods \cite{dang2014discriminative,Keller2012HiCS,noto2012frac,lazarevic2005feature}.
Consequently, they may retain features that do not optimally serve for the outlier scoring method and the performance of the subsequent outlier detection may be sufficiently biased. 
The recent work in \cite{pang2018sparse} involves the outlier scoring methods when searching the relevant feature subset.
It builds the sequential ensembles to refine feature selection and outlier scoring by iterative sparse modeling with outlier scores as the pseudo target feature. 
Though demonstrating promising performance, we observe that the used outlier detector and feature selection method are still performed in an iterative manner, which may lead to a suboptimal solution.
A question naturally raised is that can we take a step forward to integrate outlier detection and feature selection into a joint framework?

To address the aforementioned issue, this paper introduces a novel outlier detection ensemble framework with embedded feature selection, termed ODEFS. 
Specifically, ODEFS uses a given outlier scoring method to compute initial outlier scores of data objects, and then defines an outlier thresholding function to identify a set of outlier candidates.
Considering that diverse outliers may have different discriminative feature subsets \cite{wang2019parallel,wang2006research}, ODEFS builds an ensemble framework to obtain multiple feature subsets by bagging.
It randomly chooses examples from both outlier candidates and the unlabeled objects. 
They are fed into the objective function which embeds feature selection into outlier detection to learn customized feature subsets for such outlier scoring methods.
The pairwise ranking loss is adopted in the objective to encourage outliers having higher ranks than the inliers.

Notice that outlier thresholding function may produce unreliable outlier candidates since the initial outlier scores are computed using all the original features.
To improve the reliability of the training set, we propose to adopt thresholded self-paced learning to simultaneously implement example selection and feature weighting. It selects "easy" examples, i.e., the ones with small loss values which are more likely to be outliers, as the training set.
After that, we design an alternate optimization algorithm with proved convergence to obtain reliable and informative feature subsets.
Finally, ODEFS applies the same given outlier detector to the data with the selected feature subsets in a weighted aggregating manner to produce a reliable outlier scoring. 

The pairwise ranking loss function involves a large number of interactive terms between the outlier examples and the unlabeled examples, leading to a high computation complexity.
To reduce the number of interactive terms, we theoretically analyze the generalization error bound of the proposed framework, which provides valuable insights into relationships between some important parameters and the detection performance.
Those insights lead to some useful practical guidance. 
For example, we find that the improvement on the error bound is quite limited by increasing the number of unlabeled examples when it is more than the number of outlier examples.
Based on this finding, we can significantly reduce the time complexity by including a moderate number of examples without decreasing the detection performance.

The main contributions of this paper are three folds.
\begin{itemize}

\item We introduce the ODEFS framework for identifying outliers in noise data. 
Different from existing methods that separate feature selection from subsequent outlier detectors, ODEFS unifies the two tasks in a joint formulation.

\item We derive a thresholded self-paced learning algorithm to eliminate the negative effect of the unreliable outlier candidates. To solve the resultant optimization problem, we design an alternate algorithm and prove its convergence.

\item We theoretically analyze the generalization error bound of the proposed framework, which provides valuable insight into the theoretical performance of the method and helps to reduce the computation complexity. 
\end{itemize}

The ODEFS framework is instantiated on one state-of-the-art distance-based method LeSiNN \cite{Pang2016LeSiNN}.
It is also worthy of mentioning that the proposed framework can be easily extended to other formulations.
Extensive empirical results on 12 real-world data sets show that ODEFS 
(\romannumeral1) reduces a large proportion of features and improves the performance of the original bare method; 
(\romannumeral2) performs substantially better and more stably than the state-of-the-art competitors; 
(\romannumeral3) has much better resilience to noisy features than its competitors;
(\romannumeral4) has linear time complexity w.r.t. data size and feature size.

\section{Related Work}

\subsection{Outlier Detection in Noisy Data}
Subspace-based methods \cite{aggarwal2005effective,muller2011statistical,Keller2012HiCS,dang2014discriminative} are popular solutions for outlier detection in noisy data. 
They search for a set of feature subspaces and use them in an ensemble framework to avoid the negative effect of noise features, but the subspace searching is often costly as it requires extensive search in identifying the feature subspaces in high-dimensional data. 
Random subspaces generation is a widely used solution to address this efficiency issue \cite{lazarevic2005feature,nguyen2010mining}, but it may include many noisy features into subspaces.

Alternatively, feature selection-based methods aim to identify optimal feature subset(s) that reveals the exceptional behaviors of outliers. 
Although feature selection has been well investigated in clustering and classification \cite{xu2016weighted,li2018feature,nie2016unsupervised}, there exists limited work on outlier detection because it is challenging to define feature relevance to outlier detection given its unsupervised nature.  
RegFS in \cite{paulheim2015decomposition,noto2012frac} defines the relevance of features by their correlation to the other features. 
The assumption is that independent features are not useful in capturing the violation in outliers.
This assumption may be invalid since some features can be strongly relevant to outlier detection but not correlated to other features.
CINFO in \cite{pang2018sparse} firstly generates outlier scores via a given outlier detector, and then feed the scores into sparse learning based supervised feature selection to choose relevant features. These two steps are iteratively performed to build a sequential ensemble outlier detection framework. 
Such an iterative manner may result in feature subset(s) that are suboptimal to the outlier detectors.

Most of the above methods generate multiple feature subsets and work in an ensemble framework.
They combine the results calculated on these feature subsets to obtain a reliable detection result. 
In recent years, there are also some other outlier ensemble learning works that construct a set of independent base models \cite{sugiyama2013rapid,Zhang2017LSHiForest,rayana2016sequential}.
Since they work on the full feature space, their performance is still largely biased by noisy features. 
It is also worth noting that there are some successful works on joint feature selection and outlier detection for categorical data \cite{pang2017learning,pang2016unsupervised}. 
Using popular unsupervised discretization methods like equal-width and equal-frequency to adopt these methods to numeric data perform poorly \cite{pang2018sparse}.
We therefore focus on comparing ODEFS with numeric data-based methods in our experiments.
There are also some works on representation learning for outlier detection \cite{pang2018learning}. 
Although feature selection can be viewed as an approximate of representation learning, the method is customized for a given outlier detection method only. Thus it is not added to the competitors.

\subsection{Self-paced Learning}
Self-paced learning \cite{kumar2010self} is motivated by the procedure of human learning: from easy to hard. 
In machine learning problems, the value of loss function is used to measure "easiness". 
How easy examples should be used for training is controlled by a threshold $\lambda$.
Formally, given training examples $\{(\mathbf{x}_1, \mathbf{y}_1), (\mathbf{x}_2, \mathbf{y}_2), \dots, (\mathbf{x}_n, \mathbf{y}_n)\}$ and learning model $f$, the self-paced learning problem is:
\begin{equation}
\fontsize{9pt}{10pt} \selectfont
\min\limits_{\mathbf{v, w}} \sum\nolimits_{i=1}^n v_i f_\mathbf{w}(\mathbf{x}_i, \mathbf{y}_i) - \lambda v_i, \;\;
\mathbf{s.t.} \;\; v_i \in \{0, 1\}
\label{eqa:self_paced}
\end{equation}
where $\mathbf{v} = [v_1, v_2, \dots, v_n]$ are binary parameters that denote the weights of examples, $\mathbf{w}$ is the learning parameters, and $-\lambda v_i$ is called self-paced regularization term. 

When $\mathbf{v}$ is given, the minimization over $\mathbf{w}$ is a weighted loss minimization problem. 
And when $\mathbf{w}$ is fixed, the optimal $v_i$ is determined by the closed form:
\begin{equation}
\fontsize{9pt}{10pt} \selectfont
{v_i} = \left\{ \begin{array}{l}
 1,\;\;{L_i} < \lambda,  \\ 
 0,\;\;otherwise. \\ 
 \end{array} \right.
\end{equation}
where $L_i$ is the loss of $\mathbf{x}_i$ and $\lambda$ increases at each iteration by step $\delta$. All examples will be added into the training set at the end of training when $\lambda$ is large enough.

Kumar et al. \cite{kumar2010self} demonstrate that self-paced learning algorithm outperforms the state-of-the-art methods for learning a latent structural SVM on several applications. To the best of our knowledge, self-paced learning has not been used in unsupervised outlier detection yet. Maybe this is because it is hard to define the loss and set the hyper-parameters of self-paced learning for unsupervised outlier detection. We fill this gap by introducing a variant version of self-paced learning into outlier detection to select reliable outlier examples.

\section{The Proposed Algorithm}

\begin{figure}[tbp]
\centering
\includegraphics[width=0.8\columnwidth]{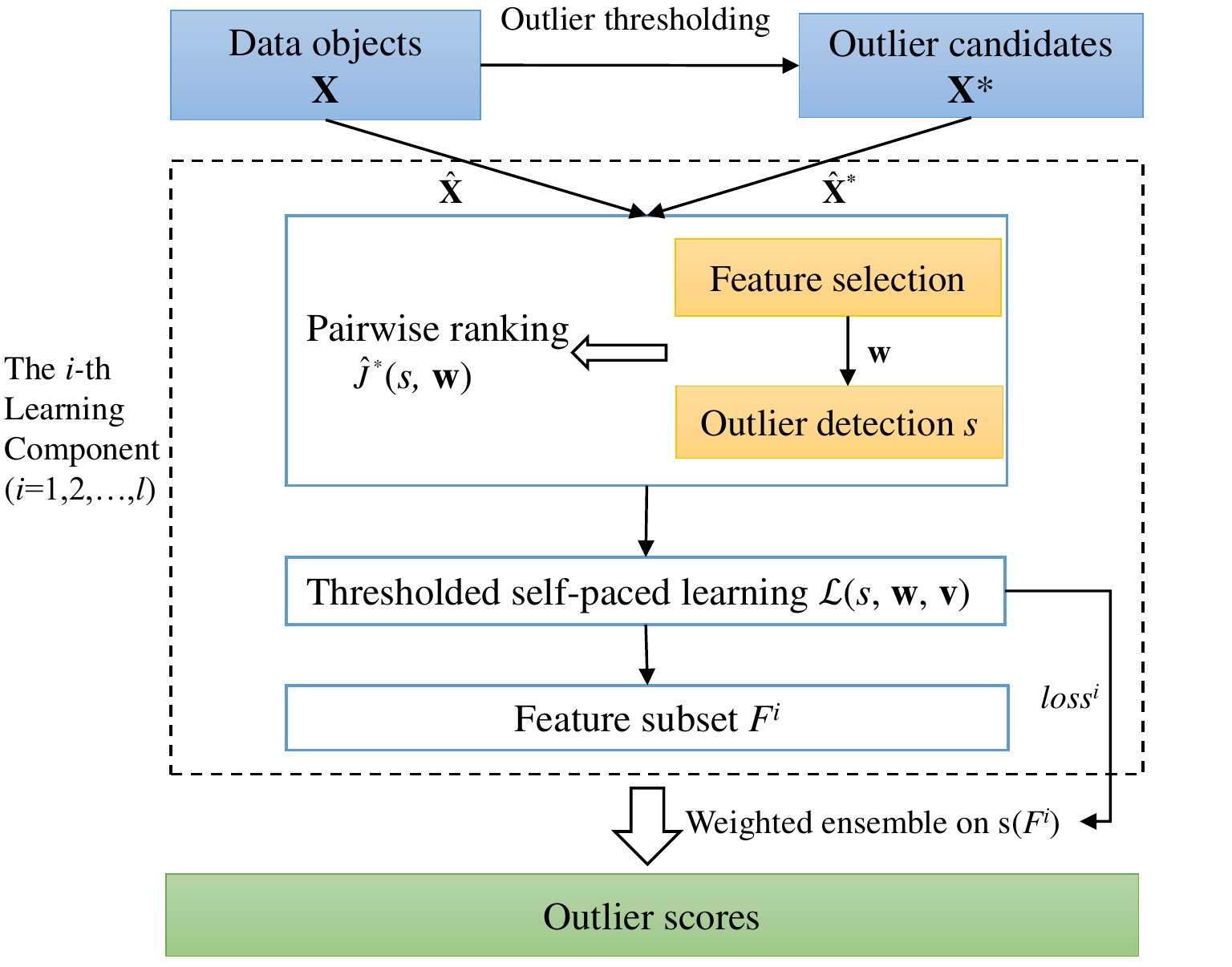}
\caption{The framework of ODEFS. ODEFS builds a parallel ensemble framework which consists of $l$ feature learning components.
It defines an outlier thresholding function to identify a set of outlier candidates $\mathbf{X}^\star$ (size $n^\star$) based on the outlier scores computed by $s$.
For each feature learning component, ODEFS randomly chooses $m$ unlabeled examples ($\hat{\mathbf{X}} = \{\hat{\mathbf{x}}_1, \hat{\mathbf{x}}_2, \dots, \hat{\mathbf{x}}_m$\}) from $\mathbf{X}$ and $m^\star$ outlier examples ($\hat{\mathbf{X}}^\star = \{\hat{\mathbf{x}}^\star_1, \hat{\mathbf{x}}^\star_2, \dots, \hat{\mathbf{x}}^\star_{m^\star}\}$) from $\mathbf{X}^\star$ .
These examples are fed into a pairwise ranking formulation that embeds feature selection into outlier detection.  
In the training process, thresholded self-paced learning is proposed to simultaneously learn example weights $\mathbf{v}$ and feature weights $\mathbf{w}$. 
With the $l$ groups of scores computed on the obtained feature subsets, ODEFS finally performs a weighted aggregating based on the learning loss to obtain the final outlier scores.}
\label{fig:framework}
\end{figure}

We consider outlier detection problems defined over a set of $n$ data objects $\mathbf{X} = \{\mathbf{x}_1, \mathbf{x}_2, \dots, \mathbf{x}_n \}$, where each data object is described as a $d$-dimensional real-valued vector $\mathbf{x}_i = \{x_{i1}, x_{i2}, \dots, x_{id} \}$. 
There is an unknown partition that divides $X$ into a set of outliers $\mathbf{X}^+ = \{\mathbf{x}_1^+, \mathbf{x}_2^+, \dots, \mathbf{x}_{n^+}^+ \}$ and a set of inliers $\mathbf{X}^- = \{\mathbf{x}_1^-, \mathbf{x}_2^-, \dots, \mathbf{x}_{n^-}^- \}$, so that $\mathbf{X} = \mathbf{X}^ + \cup \mathbf{X}^ -$.
$n^+$ and $n^-$ are the number of outliers and inliers, respectively. 
$\pi = n^+ / n$ is the outlier percentage of the data set.
Outlier detector $s(\cdot):\mathbf{x}_i \to \mathbb{R}$ assigns outlier scores to objects in $\mathbf{X}$ to yield an overall outlier ranking, with the goal of having the outliers to be higher ranked than the inliers.
We will assume that the associated outlier detector has a particular form that the feature weights can be embedded into the scoring function: $s(\mathbf{x}) \to s(\mathbf{x}, \mathbf{w})$.

The framework of ODEFS is illustrated in Figure \ref{fig:framework}, 
ODEFS tries to encourage $s(\mathbf{x}^ \star , \mathbf{w})$ to be larger than $s({\mathbf{x}^ -, \mathbf{w}})$.
Since the feature selection is guided by $s$, the chosen features only attain the information that is the most important to distinguish outliers from inliers. 
The feature subsets obtained by ODEFS are therefore tailored for the outlier detector.
It is a chicken-and-egg problem because we do not know which part of the outlier candidates are true outliers at the beginning. 
Fortunately, we can tackle this problem by iteratively selecting reliable examples according to the learning loss value in thresholded self-paced learning.
This learning strategy enables ODEFS to get more reliable discriminative feature subsets.
We detail the key steps of ODEFS in the following.

\subsection{Outlier Thresholding with Cantelli's Inequality}
The outlier thresholding function is to identify a set of most likely outliers.
We adopt the outlier thresholding function proposed in \cite{pang2018sparse,pang2018learning} which is based on Cantelli's inequality to obtain the outlier candidates: 
\begin{equation}
\fontsize{9pt}{10pt} \selectfont
\mathbf{X}^\star = \{\mathbf{x} | s(\mathbf{x}) - \mu - a \sigma > 0, x \in \mathbf{X} \}
\label{eqa:candidate_get}
\end{equation}
where $\mu$ and $\sigma^2$ are the average value and variance of all the initial outlier scores computed by $s$ with all features, and $a \ge 0$ is user-defined thresholding rate based on a desired false positive bound.

The reason we adopt Eq. (3) as initial outlier threshold is two folds: (\romannumeral1) it provides an upper bound which can be used to study the theoretical performances of the proposed model (see our theoretical foundation); (\romannumeral2) it is simple but useful as shown in the experiments.

\subsection{Pairwise Ranking Loss for Outlier Detection with Embedded Feature Selection}
Outlier scoring method $s$ yields an overall outlier ranking, with the goal of having the outliers to be higher ranked than the inliers. It tries to maximize: 
\begin{equation}
\fontsize{9pt}{10pt} \selectfont
J(s) =  \frac{1}{{n^ + }{n^ - }} \sum\nolimits_{i=1}^{n^+} \sum\nolimits_{j=1}^{n^-} {\phi (s({\mathbf{x}^ +_i }) \ge s({\mathbf{x}^ -_j }))}
\end{equation}
where $\phi$ is an indicator function that returns $1$ if the condition satisfies and 0 otherwise.

Since ODEFS is an unsupervised framework, it is impossible to directly obtain labels for both outliers and inliers. 
Inspired by \cite{ren2018robust}, we propose a relaxed pairwise ranking:
\begin{equation}
\fontsize{9pt}{10pt} \selectfont
\begin{split}
J'(s)  =  &\frac{1}{{n^ + }{n}} \sum\nolimits_{i=1}^{n^+} \sum\nolimits_{j=1}^{n} {\phi (s({\mathbf{x}^ +_i }) \ge s({\mathbf{x}_j }))} \\
 = & \frac{1}{{n^ + }{n}} ( \sum\nolimits_{i=1}^{n^+} \sum\nolimits_{j=1}^{n^-} {\phi (s({\mathbf{x}^ +_i }) \ge s({\mathbf{x}^ -_j }))} + \\
   & \sum\nolimits_{i=1}^{n^+} \sum\nolimits_{j=1}^{n^+} {\phi (s({\mathbf{x}^ +_i }) \ge s({\mathbf{x}^ +_j }))}) \\
 = & (1 - \pi)J(s) + \frac{n^+ + 1}{2 n}
\end{split}
\end{equation}
here the whole unlabeled object set $\mathbf{X}$ is used instead of $\mathbf{X}^-$. 
The above equation indicates that $J'(s)$ depends on $J(s)$ linearly.
That is, maximizing $J'(s)$ essentially maximizes $J(s)$. 
We therefore consider $J'(s)$ in the objective function rather than $J(s)$.
To approximate $\mathbf{X}^+$, we use outlier candidates set $\mathbf{X}^\star$ in the objective function.
Further, as ODEFS is a random sub-sampling based ensemble framework, each feature learning component feeds the randomly chosen examples (i.e., $\hat{\mathbf{X}}$ and $\hat{\mathbf{X}}^\star$) into the objective function:
\begin{equation}
\fontsize{9pt}{10pt} \selectfont
\begin{split}
\max \hat{J}^\star(s) = \frac{1}{m^\star m} \sum\nolimits_{i=1}^{m^\star} \sum\nolimits_{j=1}^{m} {\phi (s(\hat{\mathbf{x}}^\star_i) \ge s(\hat{\mathbf{x}}_j))}
\end{split}
\end{equation}

To put feature selection in the objective function, we embed feature weights $\mathbf{w}=\{w_i\}_{i=1}^d$ in which $w_i$ denotes the weight of $i$th feature and add sparsity constraints (i.e., $l_1-$norm) on them.
Since the indicator function $\phi(\cdot)$ is not continuous, the common treatment is to use convex and continuous surrogate function $h$ to approximate it. There are several loss functions that can be used here, such as sigmoid loss, hinge loss and logistic loss function. 
Without loss of generality, here, we focus on the logistic loss, which is defined as: $h(x) = \frac{1}{1 + exp(-x)}$. 
Then we get the new objective function:
\begin{equation}
\fontsize{9pt}{10pt} \selectfont
\begin{split}
\min\limits_{\mathbf{w}} \frac{1}{m^ \star}  \sum\nolimits_{i=1}^{m^\star} L_\mathbf{w}(\hat{\mathbf{x}}^\star_i) + \theta l_1(\mathbf{w})
\end{split}
\label{eqa:log_loss}
\end{equation}
where $L_\mathbf{w}(\hat{\mathbf{x}}^\star_i) = \frac{1}{m}\sum\nolimits_{j=1}^m \frac{1}{1 + \exp(s(\hat{\mathbf{x}}^\star_i, \mathbf{w})-s(\hat{\mathbf{x}}_j, \mathbf{w}))}$ is the loss of $\hat{\mathbf{x}}^\star_i$, $\theta=10^{-4}$ is a small constant.

\subsection{Thresholded Self-paced Learning}
Since the scores in the outlier thresholding are calculated using all the features, the outlier candidates may be not reliable. 
Here we use proposed thresholded self-paced learning to select the most confident examples. 
We combine (\ref{eqa:self_paced}) into (\ref{eqa:log_loss}) to get the final objective:
\begin{equation}
\fontsize{9pt}{10pt} \selectfont
\centering
\begin{split}
& \min\limits_{\mathbf{w},\mathbf{v}} \mathcal{L} =  \frac{1}{m^ \star}  \sum\nolimits_{i=1}^{m^\star} (v_i L_\mathbf{w}(\hat{\mathbf{x}}^\star_i) - \lambda v_i) + \theta l_1(\mathbf{w})   \\
& \;\;\;\;\;\;\;\;\;\;\;\;\;\;\;\;\;\;  \mathbf{s.t.} \;\; v_i \in \{0, 1\}
\end{split}
\label{eqa:final_loss}
\end{equation}
where $\mathbf{v} = [v_1, v_2, \dots, v_{m^\star}]^\top$ are weights of training examples, and $\lambda$ is the age parameter which controls the number
of selected examples.

In traditional self-paced learning, there are two hyper-parameters: the age parameter $\lambda$ for controlling the learning pace and step size $\delta$ for increasing $\lambda$. 
$\lambda$ increases by a step $\delta$ every several iterations. 
All examples should be added to the training set at the end of training when $\lambda$ is large enough.
But in our problem, not all the outlier candidates are reliable.
According to the definition of the loss function, the examples with lower losses value are more likely to be true outliers than the ones with higher losses.
Therefore we shall prevent the self-paced learning from selecting examples with high loss values, even at the end of the training.

We propose to constrain $\lambda$ according to the statistics of losses during the training:
\begin{equation}
\fontsize{9pt}{10pt} \selectfont
\lambda^t = \left\{\begin{array}{l}
\mu(L_{\mathbf{w}^{t-1}}) + \sigma(L_{\mathbf{w}^{t-1}}),\;\;\;\;\;\;\;\;\;\;\;\;\;\;\;\;\;\;t=1, \\ 
\max\{\lambda^{t-1}, \mu(L_{\mathbf{w}^{t-1}}) + \sigma(L_{\mathbf{w}^{t-1}})\},t>1.\\ 

\end{array} \right.
\label{eqa:lambda_compute}
\end{equation}
where $L_{\mathbf{w}^{t-1}}$ denotes losses for all examples in the $(t-1)$th iteration, $\mu(\cdot)$ and $\sigma^2(\cdot)$ are average value and variance of losses.

$\lambda$ now is thresholded by the changing losses of examples (i.e., $\lambda \le \max\limits_{t}\mu(L_{\mathbf{w}^{t}}) + \sigma(L_{\mathbf{w}^{t}})$), while it keeps a non-decreasing trend as the traditional version.
Examples with high loss are filtered by this setting.
Besides, this setting also ensures that at least half example are fed into the training process according to the Cantelli's inequality.

\subsection{Final Outlier Scoring}
Using a single component may produce high detection errors when diverse outliers may have diverse discriminative feature subsets.
We therefore further aggregate a set of sub-sampling based detection results to address this issue. 

With the $l$ groups of sub-samples involved in the training, we obtain a set of $l$ feature weight vectors $\{\mathbf{w}^1, \mathbf{w}^2, \dots, \mathbf{w}^l\}$, and their associated loss $\{loss^1, loss^2, \dots, loss^l\}$ as defined in Eq. (\ref{eqa:final_loss}).
We follow the literature \cite{nie2016unsupervised,guo2018dependence} to select the features with weights larger than a given threshold:  
\begin{equation}
\fontsize{9pt}{10pt} \selectfont
F^j = \{f_i| f_i \in F, \frac{w^j_i}{\max(\mathbf{w}^j)} > \epsilon\}
\label{eqa:feature_selection}
\end{equation}
where $\epsilon = 0.05$ is a small constant, and $\max(\mathbf{w}^j)$ denotes the maximum value in $\mathbf{w}^j$.

After that, we borrow the idea of boosting \cite{freund1997decision} to combine the outlier score vectors with associated loss as weights, and define the final outlier score for each data object in the ensemble as follows:
\begin{equation}
\fontsize{9pt}{10pt} \selectfont
final\_score(\mathbf{x}) = \sum\nolimits_{i=1}^l u^i \tau(s(\mathbf{x}, F^i))
\label{eqa:final_score}
\end{equation}
where $u^i$ is a normalized weight by $u^i = \frac{\exp(- loss^i)}{ \sum\nolimits_{j=1}^l \exp(- loss^i) }$, $s(\mathbf{x}, F^i)$ denotes outlier scoring with $F^i$, and $\tau(s(\mathbf{x}, F^i)) = \frac{s(\mathbf{x}, F^i)}{\sum\nolimits_{j=1}^n s(\mathbf{x}_j, F^i)}$ is a vector normalization function that normalizes the vector into an unit norm to address the heterogeneity of the outlier scores from heterogeneous feature subsets.

\noindent\subsection{Optimization and Convergence Analysis}
There are two parameters in Eq. (\ref{eqa:final_loss}), $\mathbf{w}$, $\mathbf{v}$, corresponding to feature learning and reliable examples selection, respectively. Motivated by the literatures \cite{kumar2010self,wang2017ta,wang2014general}, we use an alternative search strategy to optimize $\mathbf{v}$ and $\mathbf{w}$.

\subsubsection{Update $\mathbf{v}$ with $\mathbf{w}$ fixed}
With $\mathbf{w}$ fixed, we first compute $\lambda$ by Eq. (\ref{eqa:lambda_compute}). Then the optimal $\mathbf{v}$ can be easily obtained in closed form:
\begin{equation}
\fontsize{9pt}{10pt} \selectfont
{v_i} = \left\{ \begin{array}{l}
 1,\;\;{L_\mathbf{w}(\hat{\mathbf{x}}^\star_i)} < \lambda,  \\ 
 0,\;\;otherwise. \\ 
 \end{array} \right.
 \label{eqa:update_v}
\end{equation}

\subsubsection{Update $\mathbf{w}$ with $\mathbf{v}$ fixed}
With $\mathbf{v}$ fixed, Eq. (\ref{eqa:final_loss}) is simplified to
\begin{equation}
\fontsize{9pt}{10pt} \selectfont
\min\limits_{\mathbf{w}} \frac{1}{m^ \star}  \sum\nolimits_{i=1}^{m^\star} v_i L_\mathbf{w}(\hat{\mathbf{x}}^\star_i) + \theta l_1(w)
\label{eqa:update_w}
\end{equation}
which is consistent with $l_1$-norm optimization problem. Thus we can resort to SGD.

Our algorithm requires an initial parameter $\mathbf{w}^0$.
Following the literature \cite{kumar2010self}, we obtained an estimate of $\mathbf{w}^0$ by initially setting $v_i = 1$ for all outlier examples.
Then it goes to the normal $1$st iteration.

\subsubsection{Convergence analysis}
In $t-$th($t \ge 1$) iteration, we first update $\lambda$ to get $\mathcal{L}_1^t$:
\begin{equation}
\fontsize{9pt}{10pt} \selectfont
\mathcal{L}_1^t = \frac{1}{m^ \star}  \sum\nolimits_{i=1}^{m^\star} v_i^{t-1} L_{\mathbf{w}^{t-1}}(\hat{\mathbf{x}}^\star_i)-\lambda^{t}v_i^{t-1} + \theta l_1(w^{t-1})
\end{equation}
thus $\mathcal{L}_1^t < \mathcal{L}^{t-1}$ as $\lambda^{t} \ge \lambda^{t-1}$.

Then we update $\mathbf{v}$ to compute $\mathcal{L}^t_2$:
\begin{equation}
\fontsize{9pt}{10pt} \selectfont
\mathcal{L}_2^t = \frac{1}{m^ \star}  \sum\nolimits_{i=1}^{m^\star} v_i^{t} L_{\mathbf{w}^{t-1}}(\hat{\mathbf{x}}^\star_i)-\lambda^{t}v_i^{t} + \theta l_1(w^{t-1})
\end{equation}
we have:
\begin{equation}
\fontsize{9pt}{10pt} \selectfont
\mathcal{L}_2^t - \mathcal{L}_1^t = \frac{1}{m^ \star}  \sum\nolimits_{i=1}^{m^\star} (v_i^{t} - v_i^{t-1})(L_{\mathbf{w}^{t-1}}(\hat{\mathbf{x}}^\star_i)-\lambda^{t}) 
\end{equation}

According the computation of $\mathbf{v}$ in Eq. (\ref{eqa:update_v}): (\romannumeral1) if $v_i^t=1$, we have $v_i^{t} - v_i^{t-1} \ge 0$ and $L_{\mathbf{w}^{t-1}}(\hat{\mathbf{x}}^\star_i)-\lambda^{t} < 0$; (\romannumeral2) if $v_i^t=0$, we have $v_i^{t} - v_i^{t-1} \le 0$ and $L_{\mathbf{w}^{t-1}}(\hat{\mathbf{x}}^\star_i)-\lambda^{t} \ge 0$. Thus, $\mathcal{L}_2^t - \mathcal{L}_1^t \le 0$.

Lastly, when we update the feature weights $\mathbf{w}$, we have a closed form solution. The objective function is guaranteed to decrease, that is $\mathcal{L}^t < \mathcal{L}_2^t$. Then we have $\mathcal{L}^t < \mathcal{L}_2^t \le \mathcal{L}_1^t \le \mathcal{L}^{t-1}$.
And $\mathcal{L} =  \frac{1}{m^ \star}  \sum\nolimits_{i=1}^{m^\star} (v_i L_\mathbf{w}(\hat{\mathbf{x}}^\star_i) - \lambda v_i) + \theta l_1(w) > -\lambda \ge -(\max\nolimits_{t}\mu(L_{\mathbf{w}^{t}}) + \sigma(L_{\mathbf{w}^{t}})) \ge -2 $, thus the convergence of the optimization is proved.

\subsection{Time Complexity Analysis}

\begin{algorithm}[t]{}
    \caption{ODEFS}
    \textbf{Input:} Data objects $\mathbf{X}$ \\
    \textbf{Output:} Outlier scores $final\_score(\mathbf{x})$ for each $\mathbf{x}$
    \begin{algorithmic}[1]
        \State Calculate the outlier scores for $\mathbf{X}$ with $s(\cdot)$;  
        \State Obtain the outlier candidates set by Eq. (\ref{eqa:candidate_get});
        \For{$i=1 \to l$}  
            \State Randomly select $m^\star$ objects $\mathbf{X}^\star$;
            \State Randomly select $m$ objects from $\mathbf{X}$;
            \State Initialize $\mathbf{w}^0$ by optimizing Eq. (\ref{eqa:update_w})  with $\mathbf{v} = \mathbf{1}$;
            \Repeat
                \State Update $\lambda$ by Eq. (\ref{eqa:lambda_compute});
                \State Update $\mathbf{v}$ by Eq. (\ref{eqa:update_v});
                \State Update $\mathbf{w}$ by optimizing Eq. (\ref{eqa:update_w});
            \Until{convergence}
            \State Select the features by Eq. (\ref{eqa:feature_selection}); 
        \EndFor \\
        \Return Outlier scores calculated by Eq. (\ref{eqa:final_score}).
    \end{algorithmic}
\end{algorithm}

The whole algorithm is summarized in Algorithm 1.
In each learning component, the main computation cost involves the alternative optimization process in which optimization of $\mathbf{w}$ is the most complex part.
Therefore we focus on calculating the complexity of optimizing $\mathbf{w}$.
The pairwise ranking loss function involves a huge number of interactive terms between outlier examples and unlabeled examples. 
Specifically, the computation complexity of $l$ learning components can be represented as $O(l mm^\star d)$.
Fortunately, from the proposed Theorem 1 in the following section, we can see that when the number of outlier examples (i.e., $m^\star$) is fixed, the marginal gain by including more unlabeled examples (i.e., increasing $m$) is decreasing. 
Thus we set $m = 6m^\star$ according to the theoretical analysis and empirical validation.
Then the total time complexity is $O(l(m^\star)^2 d)$.
$m^\star$ is a given parameter, the ensemble size $l = 2 \lceil \frac{n^\star }{m^\star} \rceil$ tries to involve at least equal number of outlier candidates into training, so the overall time complexity can also be represented as $O(d n^\star)$. 
Since $n^\star$ is linear to $n$, the proposed ODEFS is linear w.r.t. data size and feature size.

\section{Theoretical Foundation}
In this section, we study the theoretical performances of the proposed ODEFS, which provides practical guidance of parameters setting.

\subsubsection{Theorem 1.}
Assume that all data objects in $X$ are i.i.d. samples, with probability at least $1-\delta$ we get the upper error bound of ODEFS with:
\begin{equation}
\fontsize{9pt}{10pt} \selectfont
\centering
\begin{split}
&  \;\;\;\;\;\; \hat{J}^\star- \mathbb{E}(\hat{J}^\star) \le O(  \sqrt {\frac{\kappa _{m }}{m }}  + \sqrt {\frac{\kappa _{m^ \star }}{m^ \star }} ) \\
& \frac{a^2}{1 + a^2} \mathbb{E}(J') + \frac{1-\pi}{2(1 + a^2)} \le \mathbb{E}(\hat{J}^\star) \le \mathbb{E}(J')
\end{split}
\label{eqa:theorem1}
\end{equation}
where $\hat{J}^\star$, $\mathbb{E}(\hat{J}^\star)$ and $\mathbb{E}(J')$ are respectively the adopted empirical loss, thresholding based expected loss, and ideal expected loss, and ${\kappa _{m'}}$ is defined as ${\kappa _{m'}} = d' \log (dm'/d' ) + \log \frac{1}{\delta}$ in which $d'$ is the number of selected features.

\subsubsection{Proof.}
To give a detailed proof, we first give a brief introduction of two lemmas that come from the references.

\noindent\textit{lemma 1.} The outlier thresholding function $\varphi(s, \mathbf{x}) = s(\mathbf{x}) - \mu - a \sigma$ has a false positive upper bound of $\frac{1}{(1+a^2)}$ \cite{pang2018learning}.

The above lemma is a variant of Cantelli's inequality, which implies that the probability that we could wrongly identify inliers as outliers is up to $\frac{1}{1+a^2}$ when we define the threshold as $\mu + a \sigma$.

\noindent\textit{lemma 2.} Assume that all data objects in $\mathbf{X}$ are i.i.d. samples, with probability at least $1-\delta$ we have \cite{ren2018robust}:
\begin{equation}
\fontsize{9pt}{10pt} \selectfont
\widehat{\mathrm{BAUC}}(s)- \mathrm{BAUC}(s) \le O(\sqrt {\frac{{{\kappa _n}}}{n}}  + \sqrt {\frac{\kappa _{n^ + }}{n^ + }} )
\label{eqa:lemma2}
\end{equation}
where $\widehat{\mathrm{BAUC}}$ and $\mathrm{BAUC}$ are defined as:
\begin{equation}
\fontsize{9pt}{10pt} \selectfont
\widehat{ \mathrm{BAUC}}(s) = \frac{1}{{n^ + }{n}} \sum\nolimits_{i=1}^{n^+} \sum\nolimits_{j=1}^{n} {\phi (s({\mathbf{x}^ + _i}) \ge s({\mathbf{x}_j}))}
\end{equation}
\begin{equation}
\fontsize{9pt}{10pt} \selectfont
\mathrm{BAUC} (s) = {{\mathbb E}_{{\mathbf{x}^ + } \in {D^ + }}}{{\mathbb E}_{{\mathbf{x} } \in {D}}}\phi (s({\mathbf{x}^ + }) \ge s({\mathbf{x}}))
\end{equation}
here $D^+$ and $D$ are respectively the distribution of outliers and the whole dataset.
And ${\kappa _{n^\prime}}$ is defined as
\[{\kappa _{n^\prime}} = d' \log (dn^\prime/d') + \log \frac{1}{\delta }\]
where $d'$ is the number of selected features.

Based on these two lemmas, we give below that ODEFS can obtain an upper error bound for its learning process as follows.

We can take the expectation of $\hat{J}^\star$ from its definition
\begin{equation}
\fontsize{9pt}{10pt} \selectfont
\begin{split}
\mathbb{E}(\hat{J}^\star) = & \mathbb{E} (\frac{1}{m^\star m} \sum\nolimits_{i=1}^{m^\star} \sum\nolimits_{j=1}^{m} {\phi (s(\hat{\mathbf{x}}^\star_i) \ge s(\hat{\mathbf{x}}_j))}) \\
= & \frac{1}{m^\star m} \mathbb{E} ( \sum\nolimits_{i=1}^{m^\star} \sum\nolimits_{j=1}^{m} {\phi (s(\hat{\mathbf{x}}^\star_i) \ge s(\hat{\mathbf{x}}_j))}) \\
= & \frac{1}{m^\star m}  ( \sum\nolimits_{i=1}^{m^\star} \sum\nolimits_{j=1}^{m} E_{\mathbf{x}^\star  \in \hat D^\star} \mathbb E_{\mathbf{x}  \in \hat D} \phi (s({\mathbf{x}^\star}) \ge s({\mathbf{x}}))) \\
= & \mathbb E_{\mathbf{x}^\star  \in \hat D^\star} \mathbb E_{\mathbf{x}  \in \hat D} \phi (s({\mathbf{x}^\star}) \ge s({\mathbf{x}})) \\
\end{split}
\end{equation}
where $\hat D^\star$ and $\hat D$ are respectively the distribution of outlier examples and unlabeled examples.

Since the objects in $\mathbf{X}$ are i.i.d. samples, then the objects in $\hat{\mathbf{X}}$ and $\hat{\mathbf{X}}^\star$ are also i.i.d. samples. 
Based on \textit{lemma 1}, we can easily have
\begin{equation}
\fontsize{9pt}{10pt} \selectfont
\hat{J}^\star - \mathbb {E}(\hat{J}^\star) \le O(  \sqrt {\frac{\kappa _{m }}{m}}  + \sqrt {\frac{\kappa _{m^ \star }}{m^ \star }} )
\end{equation}
holds with probability at least $1-\delta$.

Then we come to the second inequation in the theorem. As $\hat{\mathbf{X}}$ and $\hat{\mathbf{X}}^\star$ are respectively randomly sampled from $\mathbf{X}$ and $\mathbf{X}^\star$, we have $\hat{D} = D$ and $\hat{D}^\star = D^\star$. Thus  
\begin{equation}
\fontsize{9pt}{10pt} \selectfont
\begin{split}
\mathbb{E}(\hat{J}^\star) = & \mathbb E_{\mathbf{x}^\star  \in \hat D^\star} \mathbb E_{\mathbf{x}  \in \hat D} \phi (s({\mathbf{x}^\star}) \ge s({\mathbf{x}})) \\
= & \mathbb E_{\mathbf{x}^\star  \in D^\star} \mathbb E_{\mathbf{x}  \in D} \phi (s({\mathbf{x}^\star}) \ge s({\mathbf{x}}))  \\
\end{split}
\end{equation}
where $D^\star$ and $D$ denote the distributions for outlier candidates and the whole dataset, respectively.

Suppose the outlier candidate set is composed of outlier set $\mathbf{X}^\star_+$ and inlier set $\mathbf{X}^\star_-$, that is, $\mathbf{X}^\star = \mathbf{X}^\star_+ \cup \mathbf{X}^\star_-$.
$n^\star_+$ and $n^\star_-$ are respectively the number of outliers and inliers in $\mathbf{X}^\star$.

\begin{equation}
\fontsize{9pt}{10pt} \selectfont
\begin{split}
& \mathbb E_{\mathbf{x}^\star  \in D^\star} \mathbb E_{\mathbf{x}  \in D} \phi (s({\mathbf{x}^\star}) \ge s({\mathbf{x}}))  \\
 = & {{\mathbb E}_{{\mathbf{x}^\star } \in {D^\star_+ }, {\mathbf{x}^\star} \in {D^\star_- }}}{{\mathbb E}_{{\mathbf{x}} \in {D}}}\phi (s({\mathbf{x}^ \star}) \ge s({\mathbf{x}}))\\
 = & p{{\mathbb E}_{{\mathbf{x}^\star_+ } \in {D^\star_+ }}}{{\mathbb E}_{{\mathbf{x}} \in {D}}}\phi (s({\mathbf{x}^\star_ + }) \ge s({\mathbf{x}}))  + \\
  & (1-p) {{\mathbb E}_{{\mathbf{x}^\star_ - } \in {D^\star_- }}}{{\mathbb E}_{{\mathbf{x}} \in {D}}}\phi (s({\mathbf{x}^\star_-}) \ge s({\mathbf{x}})) + \\
 = & p{{\mathbb E}_{{\mathbf{x}^\star_+ } \in {D^\star_+ }}}{{\mathbb E}_{{\mathbf{x}} \in {D}}}\phi (s({\mathbf{x}^\star_ + }) \ge s({\mathbf{x}}))  + \\
  & (1-p)(1 -\pi) {{\mathbb E}_{{\mathbf{x}^\star_ - } \in {D^\star_- }}}{{\mathbb E}_{{\mathbf{x}^-} \in {D^-}}}\phi (s({\mathbf{x}^\star_-}) \ge s({\mathbf{x}^-})) + \\
  & (1-p) \pi {{\mathbb E}_{{\mathbf{x}^\star_ - } \in {D^\star_- }}}{{\mathbb E}_{{\mathbf{x}^+} \in {D^+}}}\phi (s({\mathbf{x}^\star_-}) \ge s({\mathbf{x}^+})) \\
\end{split}
\end{equation}
where $D^+$ and $D^-$ denote the distributions for the outliers and inliers, respectively.
$D^\star_+$ and $D^\star_-$ denote the distributions for the outliers and inliers in outlier candidates, respectively.
$p=n^\star_+/n^\star$ is the outlier percentage in $\mathbf{X}^\star$.
The term ${{\mathbb E}_{{\mathbf{x}^\star_ - } \in {D^\star_- }}}{{\mathbb E}_{{\mathbf{x}^ - } \in {D^ - }}}\phi (s({\mathbf{x}^ \star_- }) \ge s({\mathbf{x}^ - }))$ is a constant, because the probability that a randomly chosen inlier is ranked higher than another randomly chosen inlier should always be $\frac{1}{2}$. So we have
\begin{equation}
\fontsize{9pt}{10pt} \selectfont
\mathbb{E}(\hat{J}^\star) = p\mathbb{E}(J') + \frac{(1-p)(1-\pi)}{2} + (1-p)\pi \Delta 
\end{equation}
where $\Delta = {{\mathbb E}_{{\mathbf{x}^\star_ - } \in {D^\star_- }}}{{\mathbb E}_{{\mathbf{x}^+} \in {D^+}}}\phi (s({\mathbf{x}^\star_-}) \ge s({\mathbf{x}^+}))$

It is reasonable to assume: (\romannumeral1) $\mathbb{E}(J') \ge \frac{1}{2}$, this is because the probability that a chosen outlier is ranked higher than a chosen unlabeled object should be larger than $\frac{1}{2}$; (\romannumeral2) $0 \le \Delta \le \frac{1}{2}$, this is because the probability that a chosen inlier is ranked higher than a chosen outlier should be smaller than $\frac{1}{2}$. 
Based on \textit{lemma 1}, we have $a^2 / (1 + a^2) \le p \le 1$. Thus we get:
\begin{equation}
\fontsize{9pt}{10pt} \selectfont
\begin{split}
\mathbb{E}(\hat{J}^\star) = & \frac{1}{2} + (\mathbb{E}(J') - \frac{1}{2})p - \frac{\pi}{2} + \frac{p\pi}{2} + (1-p)\pi \Delta\\
  \ge & \frac{1}{2} + (\mathbb{E}(J')  - \frac{1}{2})*a^2 / (1 + a^2) - \frac{\pi}{2} + \frac{\pi a^2}{2(1+a^2)} \\ 
   =  & \frac{a^2}{1 + a^2} \mathbb{E}(J') + \frac{1-\pi}{2(1 + a^2)}
\end{split}
\end{equation}
and
\begin{equation}
\fontsize{9pt}{10pt} \selectfont
\begin{split}
\mathbb{E}(\hat{J}^\star) \le & p\mathbb{E}(J') + \frac{(1-p)(1-\pi)}{2} + \frac{(1-p)\pi}{2} \\
\le & p\mathbb{E}(J') + \frac{1-p}{2} \\
\le & p\mathbb{E}(J') + (1-p)\mathbb{E}(J') \\
= & \mathbb{E}(J')
\end{split}
\end{equation}

The proof is completed.

This theorem provides the upper bound of the difference between $\hat{J}^\star$ and the ideal expected $\mathbb{E}(J')$. 
It contains three parameters: the thresholding rate $a$, the number of sampled outliers $m^\star$ and the number of sampled unlabeled objects $m$.
This leads to the following interesting observations: 

(\romannumeral1) On one hand, $\mathbb{E}(\hat{J}^\star)$ gets close to $\mathbb{E}(J')$  when $a$ increases; on the other hand, increasing $a$ will decrease $n^\star$, thus the number of outlier candidates is limited;

(\romannumeral2) When $m^\star$ is fixed, the improvement on this bound by increasing $m$ is quite limited. Thus it is not necessary to set a large value for $m$.

According to these observations, we therefore set the parameters $a = 2$ (i.e., $\mathbb{E}(\hat{J}^\star) > 0.8\mathbb{E}(J')$ ) and $m = 6m^\star$ in our experiments. The experimental results in the following section have demonstrated the effectiveness of the parameters settings.

\section{Experimental Evaluation}
\begin{table*}[htbp]
\caption{The details of 12 used datasets, the performance of bare LeSiNN (denoted by AUC, p@k) and ODEFS-enabled LeSiNN (denoted by AUC', p@k').
d' is the average number of features retained by ODEFS-enabled LeSiNN.}
\centering
    \begin{tabular}{|c|ccccc|cc|ccc|}
    \hline\hline
    Data       & n          & d          & on         & or         & domain     & AUC    & p@k   & d'         & AUC'   & p@k' \\ \hline\hline
    Advertisements  & 3279       & 1555       & 454        & 0.138      & medical    & 0.722      & 0.445      & 48         & 0.906      & 0.674 \\ \hline
    AID362          & 4279       & 114        & 60         & 0.014      & medical    & 0.662      & 0.017      & 32         & 0.671      & 0.050 \\ \hline
    aPascal         & 12695      & 64         & 176        & 0.014      & medical    & 0.750      & 0.000      & 14         & 0.889      & 0.000 \\ \hline
    Bank            & 41188      & 53         & 4640       & 0.113      & social     & 0.597      & 0.190      & 18         & 0.641      & 0.275 \\ \hline
    Probe           & 64759      & 67         & 4166       & 0.064      & security   & 0.958      & 0.775      & 10         & 0.965      & 0.879 \\ \hline
    U2R             & 60821      & 40         & 228        & 0.004      & security   & 0.988      & 0.592      & 9          & 0.991      & 0.618 \\ \hline
    Arrhythmia      & 452        & 256        & 66         & 0.146      & nature     & 0.781      & 0.500      & 15         & 0.814      & 0.515 \\ \hline
    Mnist           & 7603       & 100        & 700        & 0.092      & nature     & 0.854      & 0.407      & 20         & 0.904      & 0.494 \\ \hline
    Musk            & 3062       & 166        & 97         & 0.032      & nature     & 1.000      & 1.000      & 24         & 1.000      & 1.000 \\ \hline
    Optdigits       & 5216       & 64         & 150        & 0.029      & nature     & 0.712      & 0.040      & 20         & 0.818      & 0.047 \\ \hline
    Speech          & 3686       & 400        & 61         & 0.017      & nature     & 0.468      & 0.016      & 58         & 0.482      & 0.049 \\ \hline
    Census          & 299285     & 503        & 18568      & 0.062      & social     & 0.602      & 0.054      & 60         & 0.701      & 0.075 \\ \hline\hline
    \end{tabular}%
\label{tab:datainfo_improvement}%
\end{table*}%

\subsection{Experiment Setup}

\subsubsection{Application to Distance-based Outlier Detection}
There are a number of outlier detectors, where different criterions are used in different algorithms. 
Here we choose one state-of-the-art distance-based outlier detector LeSiNN \cite{Pang2016LeSiNN} as representative to motivate our model. 
It is worthy mentioning that we also get similar results on another tree-based outlier detector iForest \cite{liu2012isolation}.

Given a dataset $\mathbf{X}$ of vector-valued objects, LeSiNN analyzes $\mathbf{X}$ to construct a set of random subsets.
The outlier score of an object $\mathbf{x}$ is assigned as the average value of its nearest distances to the subsets. 
\begin{equation}
\fontsize{9pt}{10pt} \selectfont
s(\mathbf{x}) = \frac{1}{c} \sum\nolimits_{i=1}^c nn\_dist_i(\mathbf{x})
\label{eqa:lesinn}
\end{equation}
where $c$ is the number of subsets and $nn\_dist_i(\mathbf{x})$ returns the nearest neighbor distance of $\mathbf{x}$ to the $i$th subset.

In particular, the weight of each feature in the distance calculation is $1$ (constant). 
It is easy to embed the feature weight into the distance calculation. 
Take Squared Euclidean Distance as an example, the weighted distance is:
$dist(\mathbf{x}_i, \mathbf{x}_j, \mathbf{w}) = \sum\nolimits_{k=1}^d w_k *(x_{ik}-x_{jk})^2$.
Thus, we can get the weighted version of LeSiNN by substituting this formulation into Eq. (\ref{eqa:lesinn}).

\subsubsection{Datasets}
As shown in Table \ref{tab:datainfo_improvement}, 12 real-world datasets are used, which cover diverse domains, i.e., medical, social, security and nature\footnote{They are available at http://archive.ics.uci.edu/ml/index.php, http://odds.cs.stonybrook.edu/, http://vision.cs.uiuc.edu/attributes/}.
They are described with four data factors, i.e., n - the number of objects, d - the number of features, on - the number of outliers and or - the outlier percentage.
Some datasets like AD, AID362, Probe, and U2R contain semantically real outliers. 
For the other datasets, we follow the literature \cite{pang2018learning,paulheim2015decomposition} to treat rare classes as outliers and the largest class as the normal class. 

\subsubsection{Parameters setting}
ODEFS and its competitors are implemented in Python 3.4.
All the experiments are executed at a PC in a 3.6GHz CPU with 16GB memory. 
In our experiments, ODEFS uses $m^\star=32$ for small datasets (i.e., $n \le 10^4$) and $m^\star=64$ for large datasets (i.e., $n > 10^4$).
Other parameters setting, i.e., $a=2$, $m = 6m^\star$, $l=2 \lceil \frac{n^\star}{m^\star} \rceil$, has been explained in the above sections.
The parameters of LeSiNN are set as the recommended settings.

\subsubsection{Evaluation Methods}
Following the literature \cite{campos2016evaluation,cheng2019neural}, we evaluate the outlier detection performance by AUC and p@k.
Their values range from $0$ to $1$ and higher value indicates better feature subset.
The Wilcoxon signed rank test is used to examine the significance of the performance of ODEFS against its competitors.  
We repeat each experiment 20 times and average the results to get a convincing evaluation.

\begin{table*}[tbp]
\fontsize{8pt}{9pt} \selectfont
  \centering
  \caption{The performance of ODEFS and its competitors. The best results are in bold. AVG is the averaged performance of a method over all datasets. p-value of Wilcoxon signed rank test is reported in the bottom.}
    \begin{tabular}{|c|cccccc|cccccc|}
    \hline\hline
            & \multicolumn{6}{c|}{AUC} &  \multicolumn{6}{c|}{p@k} \\\hline
    Data         & RandFS         & DisFS      & RegFS      & CINFO   & ODEFS$^*$   & ODEFS   & RandFS         & DisFS      & RegFS      & CINFO   & ODEFS$^*$   & ODEFS      \\\hline\hline
    Advertisements & 0.735      & 0.742      & 0.747      & 0.856          &0.832 & \textbf{0.906 } & 0.511      & 0.522      & 0.535      & 0.604      & 0.588 & \textbf{0.674 }              \\\hline
    AID362         & 0.654      & 0.648      & 0.652      & 0.663          &0.649 & \textbf{0.671 } & 0.017      & 0.017      & 0.017      & 0.033      & 0.017 & \textbf{0.050 }              \\\hline
    aPascal        & 0.742      & 0.736      & 0.752      & 0.834          &0.835 & \textbf{0.889 } & 0.000      & 0.000      & 0.000      & 0.000      & 0.000 & \textbf{0.000 }              \\\hline
    Bank           & 0.592      & 0.604      & 0.605      & 0.607          &0.601 & \textbf{0.641 } & 0.189      & 0.192      & 0.202      & 0.228      & 0.247 & \textbf{0.275 }              \\\hline
    Probe          & 0.960      & 0.951      & 0.958      & 0.958          &0.965 & \textbf{0.965 } & 0.774      & 0.776      & 0.768      & 0.773      & 0.776 & \textbf{0.879 }              \\\hline
    U2R            & \textbf{0.995 } & 0.972      & 0.975      & 0.989     &0.992 & 0.991      & \textbf{0.627 } & 0.592      & 0.592      & 0.610      & 0.605 & 0.618                   \\\hline
    Arrhythmia     & 0.792      & 0.765      & 0.732      & 0.796          &0.801 & \textbf{0.814 } & 0.500      & 0.470      & 0.439      & 0.455      & 0.470 & \textbf{0.515 }              \\\hline
    Mnist          & 0.865      & 0.865      & 0.842      & 0.875          &0.882 & \textbf{0.904 } & 0.416      & 0.419      & 0.409      & 0.427      & 0.446 & \textbf{0.494 }              \\\hline
    Musk           & 1.000      & 1.000      & 1.000      & 1.000          &1.000 & \textbf{1.000 } & 1.000      & 1.000      & 1.000      & 1.000      & 1.000 & \textbf{1.000 }              \\\hline
    Optdigits      & 0.722      & 0.721      & 0.731      & 0.681          &0.712 & \textbf{0.818 } & 0.047      & 0.047      & 0.033      & 0.027      & 0.033 & \textbf{0.047 }              \\\hline
    Speech         & 0.472      & 0.465      & 0.481      & \textbf{0.485} &0.480 & 0.482      & 0.033      & 0.033      & 0.033      & 0.049           & 0.033 & \textbf{0.049 }              \\\hline
    Census         & 0.638      & 0.642      & 0.635      & 0.642          &0.651 & \textbf{0.701 } & 0.059      & 0.059      & 0.061      & 0.067      & 0.068 & \textbf{0.075 }              \\\hline \hline
    Average          & 0.764      & 0.759      & 0.759      & 0.782        &0.783 & \textbf{0.815 } & 0.348      & 0.344      & 0.341      & 0.356      & 0.357 & \textbf{0.390 }               \\\hline
    p-value      & 0.004      & 0.003      & 0.003      & 0.006      & 0.007 &            & 0.011      & 0.008      & 0.005      & 0.008      & 0.005 &                               \\ \hline \hline
    \end{tabular}%
  \label{tab:auc_comparison}%
\end{table*}%

\subsection{Empirical Validation of Theorem 1}
\subsubsection{Experimental setting}
This section conducts empirical experiments to study how the number of unlabeled examples affects AUC.
We follow the literature \cite{zimek2012survey} to create a 100-dimensional synthetic dataset of size 10000 with 20 relevant features. 
Inliers are from a Gaussian distribution $\mathcal N(1,0.2)$, while outliers are from another Gaussian distribution $\mathcal N(1.2,0.2)$ in relevant features, and the other features are from a same Gaussian distribution $\mathcal N(1,0.2)$ and used as noisy features. 
The number of outlier examples is fixed as $m^\star=32$. We gradually increase the number of unlabeled examples from $m=m^\star$ to $m=12m^\star$ by a step $m^\star$.

\begin{figure}[htbp]
\centering
\includegraphics[width=0.6\columnwidth]{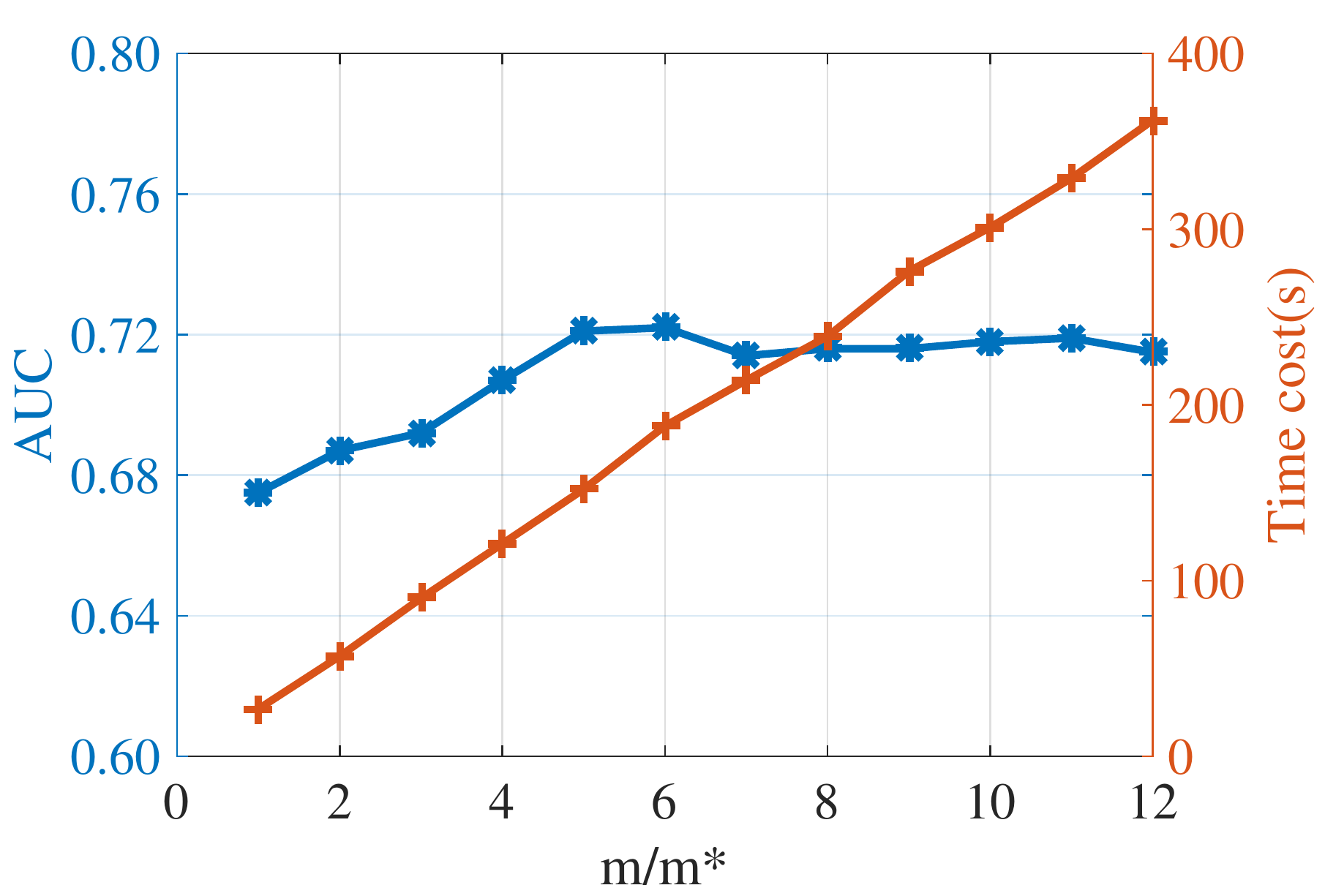}
\caption{The trends of AUC and runtime regarding different sizes of the unlabeled examples. After $m/m^\star$ is larger than 6, AUC stays stable.}
\label{fig:fig_lenn}
\end{figure}

\subsubsection{Results}
According to the experimental results shown in Figure \ref{fig:fig_lenn}, it is observed that:
It indicates that when the number of unlabeled examples is more than 6 times of the number of outlier examples, the improvement on AUC becomes quite minor.
In the meanwhile, the runtime grows linearly w.r.t the size of the examples. 
This observation is consistent with our analysis in Theorem 1. 
It essentially suggests that it is not necessary to include all the unlabeled examples in the training process when the unlabeled data points are substantially more than the outlier examples.

\subsection{Improvement to The Bare Method}

\subsubsection{Experimental Setting}
We compare the ODEFS-enabled LeSiNN with its bare version to evaluate whether ODEFS can remove noisy features and improve the performance.

\subsubsection{Results}
Table \ref{tab:datainfo_improvement} shows the feature reduction and detection performance of ODEFS-enabled LeSiNN, compared to LeSiNN performing in the original feature space. 
ODEFS-enabled LeSiNN works with only $5\%$ (e.g., on Advertisements) to less than $50\%$ (e.g., on Optdigits) of the original features, while its performance is substantially better than, or roughly the same as, its bare version. 
ODEFS enables LeSiNN to gain more than $7\%$ and $16\%$ improvement on average in terms of AUC and p@k, respectively. 
Our significance test shows that ODEFS enables LeSiNN to achieve significantly better performance at the $95\%$ confidence level.

ODEFS embeds feature learning into outlier scoring by a joint framework, which enables ODEFS to safely remove noisy features in these datasets. 
As a result, ODEFS-enabled LeSiNN works on much cleaner datasets and thus can achieve significant performance improvement.

\subsection{Comparing to State-of-the-art Methods}

\subsubsection{Experimental Settings}
Four state-of-the-art methods are used as competitors, they are: RandFS \cite{lazarevic2005feature}, RegFS \cite{paulheim2015decomposition}, DisFS \cite{dang2014discriminative}, and CINFO \cite{pang2018sparse} from four different but relevant researches.
Besides, one variant of ODEFS (named ODEFS$^*$) is also evaluated to show the contribution of self-paced learning.

\begin{itemize}

\item RandFS: RandFS is a random subspace-based method in which the features are randomly selected;

\item RegFS: RegFS is a relevance analysis based method which returns a feature relevance ranking and selects the top-ranked features;

\item DisFS: DisFS is chosen as a representative subspace-based algorithm that uncovers outliers in subspaces of reduced dimensionality in which they are well discriminated from regular objects;

\item CINFO: CINFO is a representative of feature-selection method that performs lasso-based sparse regression by treating the outlier scores as the targets to obtain feature subsets. It iteratively refines their performance by sequential ensemble; 

\item ODEFS$^*$: ODEFS$^*$ is a variant of ODEFS in which self-paced learning is removed. It feeds all the outlier examples into the training process. 

\end{itemize}

\subsubsection{Results}
Table \ref{tab:auc_comparison} shows the performance of LeSiNN with each method over all datasets.
According to the experimental results, ODEFS gets the best performance on ten and eleven of twelve datasets in terms of AUC and p@k respectively, while the performance on the other ones is close to the best.
ODEFS averagely performs better than four competitors RandFS, RegFS, DisFS, CINFO and ODEFS$^*$ by $4\%-7\%$ in term of AUC.
And in term of p@k, the average improvements are $9\%-15\%$.
The small p-values show that the improvement is significant at a high confidence level (i.e., $95\%$).

Different from RandFS, RegFS and DisFS that ignore the outlier scoring methods when they perform feature selection, ODEFS couples these two tasks in a joint formulation. 
This enables ODEFS to substantially reduce its detection errors and obtain significant AUC improvement, especially in noisy datasets like Advertisements, aPascal, and Census, which likely contain a large proportion of noisy features.
Since there is no thresholded self-paced learning in ODEFS$^*$, the quality of the training set may be reduced, leading to worse detection performance.

ODEFS and CINFO are two different feature selection based methods. 
CINFO iteratively performs lasso-based sparse regression by treating the outlier scores as the target and the original features as the predictors on the outlier candidates to obtain a few feature subsets.
The outlier scoring method and lasso-based feature selection are still in a sequential order. 
And it adopts all the outlier candidates which may contain inliers, leading to a poor feature learning performance.
In contrast, ODEFS works in a unified framework, which embeds feature selection into outlier scoring method. 
Besides, thresholded self-paced learning is proposed to improve the quality of the training set.
Thus ODEFS has better performance than CINFO.

\subsection{Capability of Handling Noisy Features}
\subsubsection{Experimental settings}
We create a few 100-dimensional synthetic datasets with different percentages of relevant features (or noisy features). 
Inliers are from a Gaussian distribution $\mathcal N(1,0.2)$, while outliers are from another Gaussian distribution $\mathcal N(1.2,0.2)$ in relevant features, and the other features are from a same Gaussian distribution $\mathcal N(1,0.2)$ and used as noisy features. 

\begin{figure}[htbp]
\centering
    \subfigure{ 
    \includegraphics[width=0.46\columnwidth]{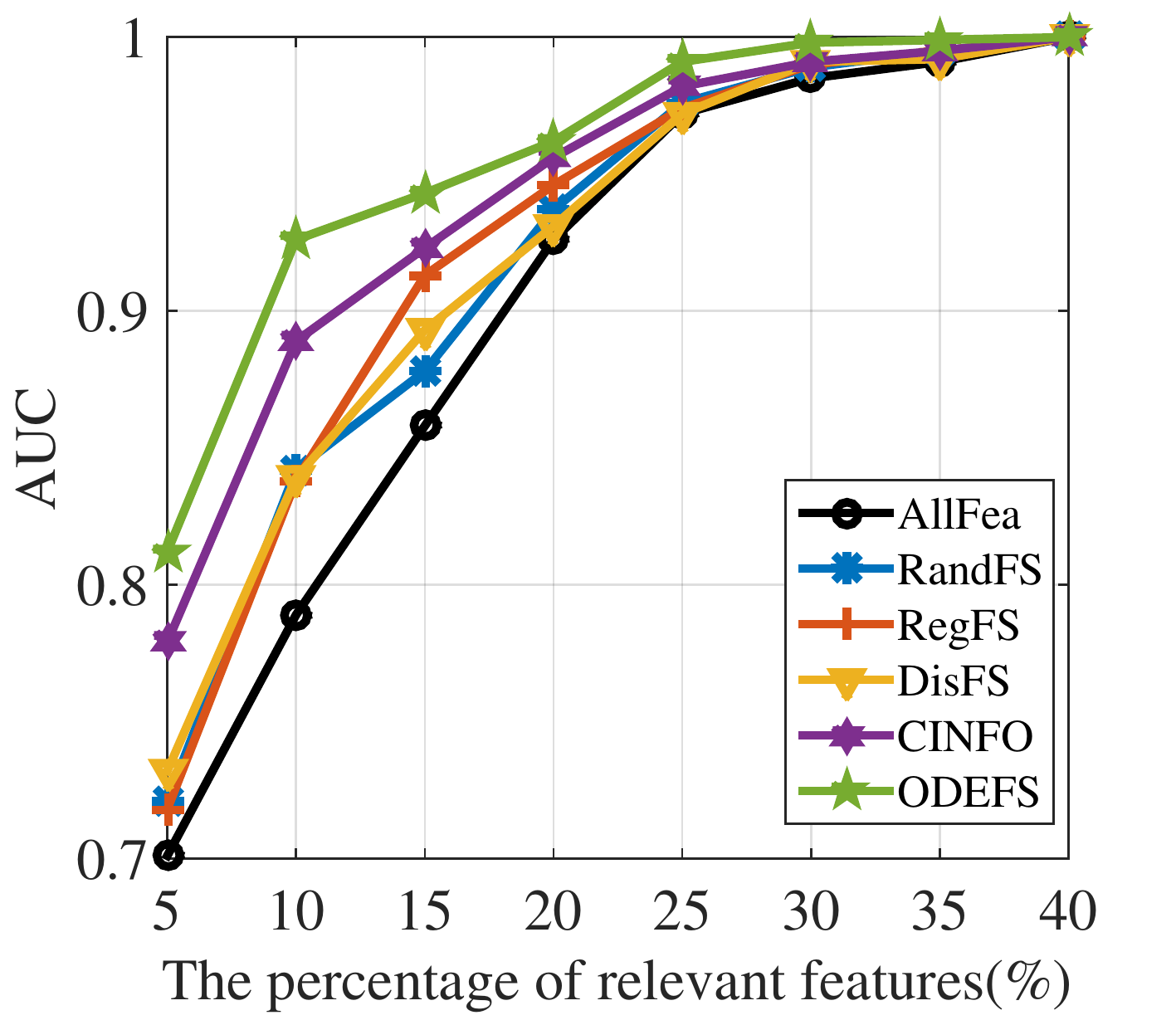}
    }
    \subfigure{
    \includegraphics[width=0.46\columnwidth]{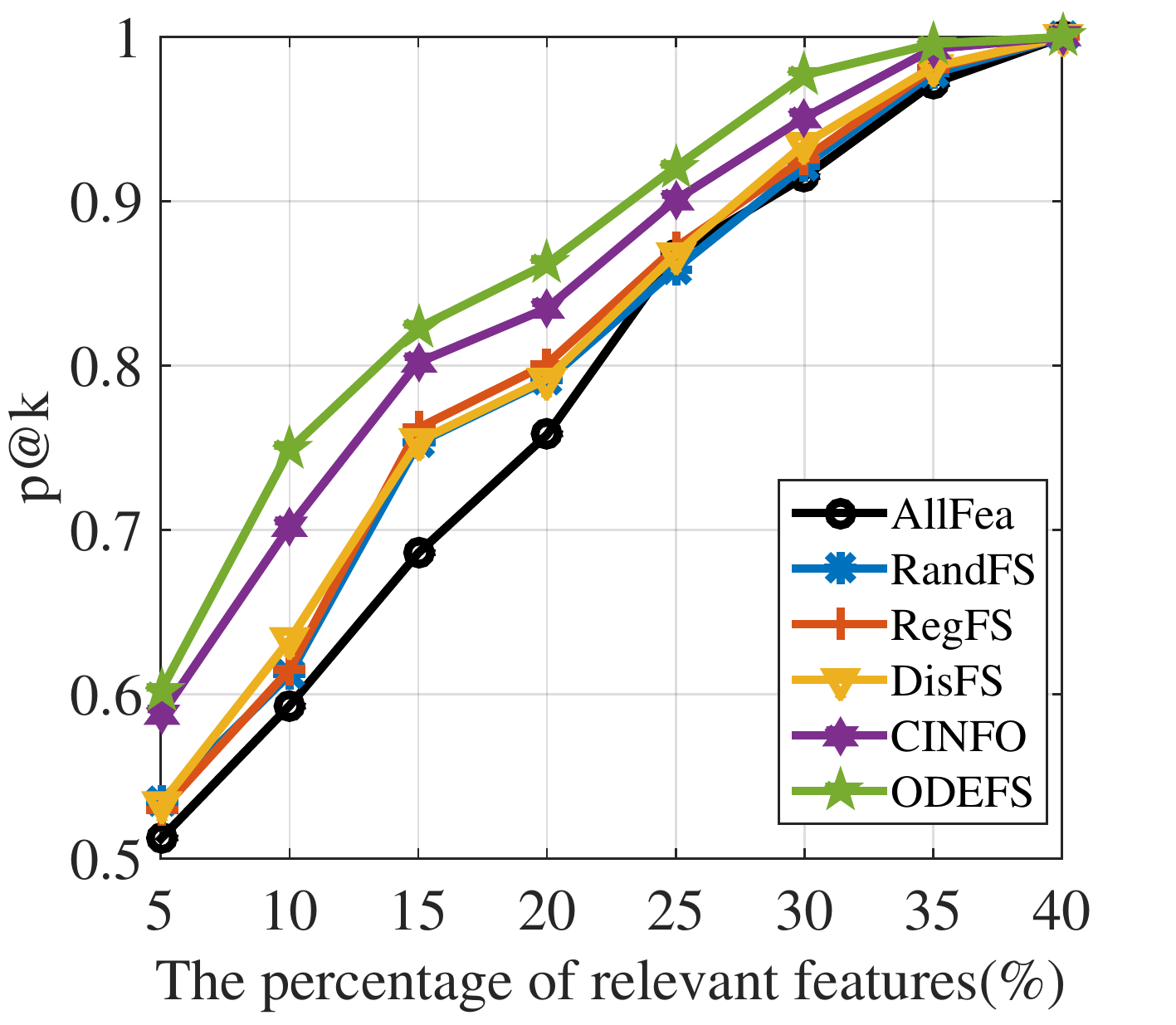}
    }
\caption{Detection performance on datasets with different levels of relevant features. ODEFS persistently performs better than its competitors. All the methods obtain AUC of nearly one with more than $35\%$ relevant features.}
\label{fig:noisy_res}
\end{figure}

\subsubsection{Results}
The performance on the synthetic datasets is shown in Figure \ref{fig:noisy_res}. 
ODEFS-enabled LeSiNN performs consistently better than five other versions in a wide range of noise levels. 
The better performance of the ODEFS-enabled LeSiNN over the competitors shows its stronger capability of handling noisy features.

\subsection{Scalability}

\subsubsection{Experimental settings}
We follow the literature \cite{pang2018sparse} to generate datasets by varying the feature size w.r.t. to a fixed data size (i.e., 1000), as well as varying the data size while fixing the feature size (i.e., 100), respectively.

\begin{figure}[htbp]
\centering
    \subfigure{ 
    \includegraphics[width=0.46\columnwidth]{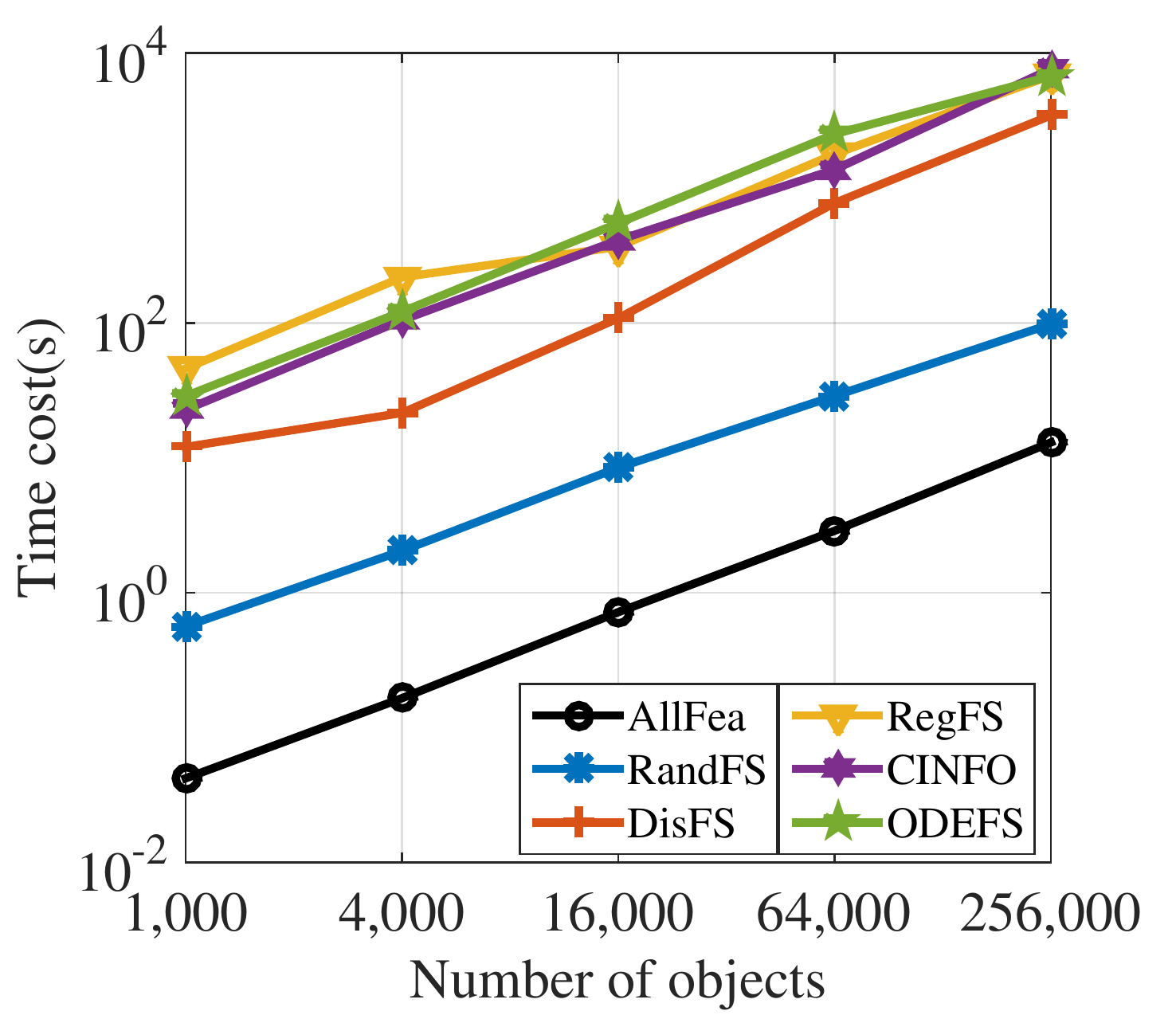}
    }
    \subfigure{
    \includegraphics[width=0.46\columnwidth]{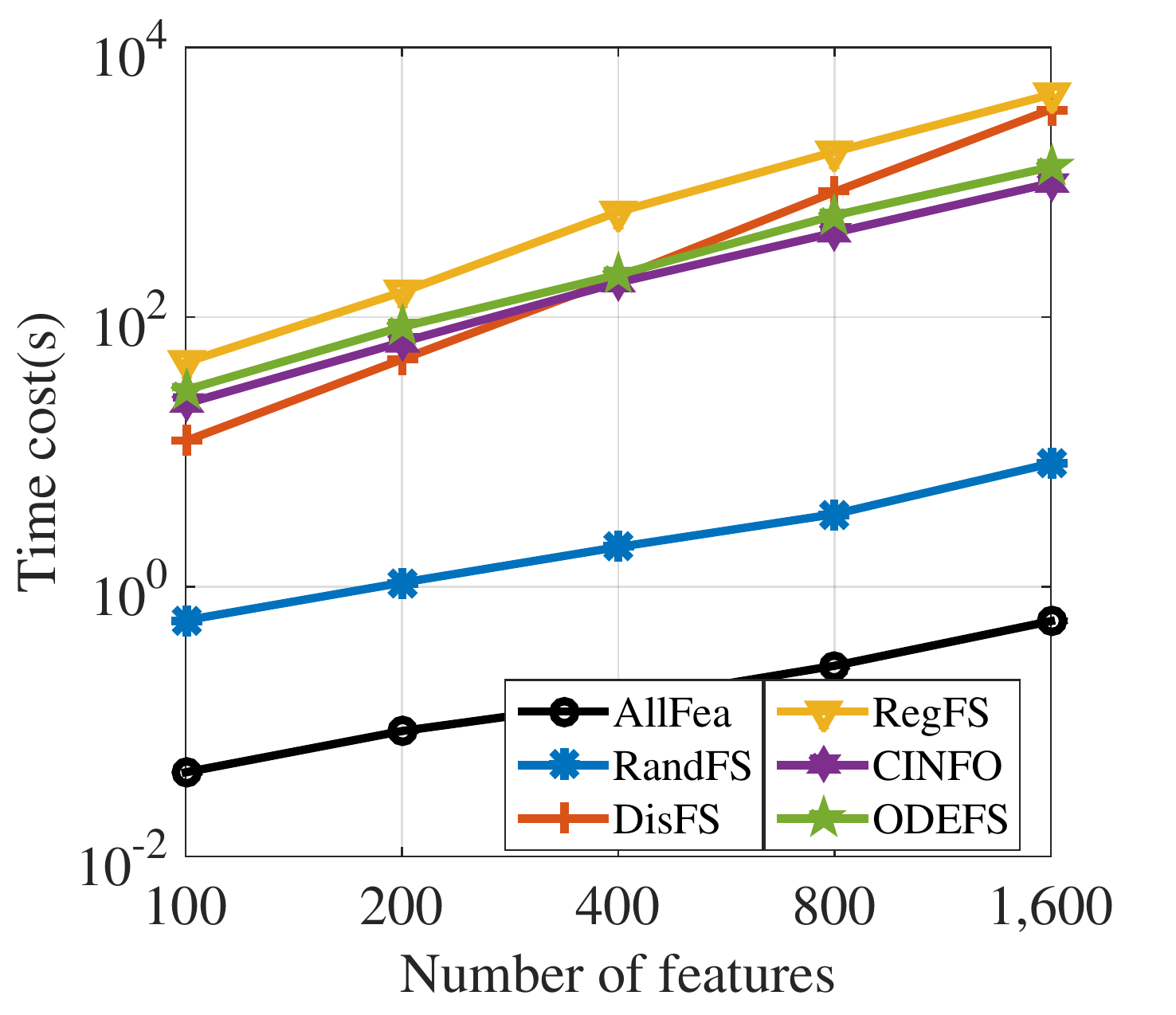}
    }
\caption{Scalability test w.r.t. data size and feature size on ODEFS and its competitors.}
\label{fig:sca_test}
\end{figure}

\subsubsection{Results}
The runtime of the six versions of LeSiNN is shown in Figure \ref{fig:sca_test}. 
ODEFS has time complexity linear w.r.t. both data size and feature size, which justifies our complexity analysis. 
In the left sub-figure, ODEFS is comparably fast to RegFS, CINFO and DisFS. 
These four methods are slower than RandFS and the bare LeSiNN, since they incorporate more sophisticated components to enhance the performance of LeSiNN. 
In the right sub-figure, RegFS is the slowest one since it has quadratic complexity while the other methods have linear time complexity.
The runtime of subspace-based method DisFS grows very fast because the subspace searching is often costly in high-dimensional data.

\section{Conclusion}
In this paper, we propose an outlier detection ensemble framework, called ODEFS, which directly embeds feature selection into outlier detection. 
We propose thresholded self-paced learning to improve the reliability of the training set and design an alternative algorithm to address the optimization problem.
Experimental results on various datasets demonstrate the effectiveness of ODEFS.

\section{Acknowledgments}
This work was supported by the National Key Research and Development Program of China (Grant No.2016YFB1000101), the National Natural Science Foundation of China (Grant No.61379052, No.61872371 and No.61872377), the Science Foundation of Ministry of Education of China (Grant No.2018A02002), the Natural Science Foundation for Distinguished Young Scholars of Hunan Province (Grant No.14JJ1026).

\bibliographystyle{aaai}
\bibliography{6641_references}

\end{document}